\documentclass[runningheads]{llncs}
\usepackage{accv}

\usepackage{accvabbrv}

\usepackage{graphicx}
\usepackage{booktabs}
\usepackage{multirow}
\usepackage{dsfont}

\usepackage{float}
\usepackage{capt-of}
\let\llncsOriginalSubparagraph\subparagraph
\let\subparagraph\paragraph
\usepackage{titlesec}
\let\subparagraph\llncsOriginalSubparagraph
\usepackage{xcolor}
\usepackage{colortbl}
\usepackage{upgreek}

\newcommand{\ind}{\mathds{1}}

\usepackage[accsupp]{axessibility}

\usepackage[
    breaklinks,
    colorlinks=true,
    citecolor=accvblue,
    linkcolor=accvblue,
    urlcolor=accvblue
]{hyperref}

\usepackage{orcidlink}

\AtBeginDocument{%
  \setlength{\abovedisplayskip}{5pt plus 1pt minus 1pt}
  \setlength{\belowdisplayskip}{5pt plus 1pt minus 1pt}
  \setlength{\abovedisplayshortskip}{3pt plus 1pt minus 1pt}
  \setlength{\belowdisplayshortskip}{3pt plus 1pt minus 1pt}
  \setlength{\jot}{2pt}
}

\titlespacing*{\section}{0pt}{14pt}{10pt}
\titlespacing*{\subsection}{0pt}{14pt}{6pt}
\titlespacing*{\subsubsection}{0pt}{12pt}{8pt}

\definecolor{multistagebg}{RGB}{242,243,245}
\definecolor{onestagebg}{RGB}{231,240,250}

\makeatletter

\renewenvironment{figure}
  {\setlength\abovecaptionskip{2\p@}%
   \setlength\belowcaptionskip{4\p@}%
   \@float{figure}}
  {\end@float}
\renewenvironment{figure*}
  {\setlength\abovecaptionskip{2\p@}%
   \setlength\belowcaptionskip{4\p@}%
   \@dblfloat{figure}}
  {\end@dblfloat}

\renewenvironment{table}
  {\setlength\abovecaptionskip{10\p@}%
   \setlength\belowcaptionskip{2\p@}%
   \@float{table}}
  {\end@float}
\renewenvironment{table*}
  {\setlength\abovecaptionskip{10\p@}%
   \setlength\belowcaptionskip{2\p@}%
   \@dblfloat{table}}
  {\end@dblfloat}

\makeatother

\begin{document}

\title{Relation-Transition-Aware Transformer\\for Video Scene Graph Generation}

\titlerunning{RTAformer}

\author{
Jaeyeon Heo\orcidlink{0009-0000-2254-1105} \and
Geuntaek Lim\orcidlink{0000-0003-2204-6109} \and
Yukyung Choi\textsuperscript{$\dagger$}\orcidlink{0000-0002-9970-0132}}

\authorrunning{J.~Heo et al.}

\institute{
Sejong University, Seoul, Republic of Korea\\
\email{\{jyheo,gtlim,ykchoi\}@rcv.sejong.ac.kr}}

\maketitle

\renewcommand\thefootnote{$\dagger$}
\footnotetext{Corresponding author.}

\begin{abstract}
\sloppy
    Video scene graph generation (VidSGG) aims to capture objects and their evolving relationships in videos.
    This requires accounting for both the persistence of object identities and abrupt changes in relationships.
    Existing methods aggregate information across frames to exploit temporal context, but an emphasis on consistency can obscure these changes.
    As a result, a model may retain its previous prediction even after the underlying relation has changed.
    We refer to this failure mode as relational semantic inertia.
    To address this issue, we propose Relation-Transition-Aware Transformer (RTAformer), a one-stage framework that separately models entity persistence and relation changes. 
    It injects calibrated differences between adjacent-frame relation queries to guide predicate updates, while cross-frame instance alignment during training helps reduce cross-frame representational drift.
    Experiments on Action Genome show improvements in recall and mean recall over existing methods, with ablation studies supporting the contributions of the proposed components and comparisons using matched backbones showing larger gains on relation-transition frames than on non-transition frames.
    Code is available at \href{https://github.com/sejong-rcv/RTAformer}{\textcolor{magenta}{https://github.com/sejong-rcv/RTAformer}}.

    
\end{abstract}

\section{Introduction}

    Video scene graph generation (VidSGG) represents objects and their relationships in videos as structured $\langle$\emph{subject, predicate, object}$\rangle$ triplets, with entities as nodes and relationships as edges.
    By explicitly representing who interacts with what and how these relationships evolve, video scene graphs provide structured information for action selection in embodied navigation and planning~\cite{huang2025esca, yin2024sgnav, rosinol2020dynamic} and for interpreting events in video-language reasoning and question answering~\cite{nguyen2025hyperglm, cherian2022spatiotemporal, urooj2023shgvqa}.
    Approaches to VidSGG have evolved from multi-stage pipelines for object detection, temporal association, and relation classification~\cite{cong2021sttran, li2022apt, nag2023tempura} to one-stage frameworks that integrate instance and relation prediction, reducing reliance on external tracking~\cite{wang2024oed, wang2025arn, le2026uno}.

\begin{figure}[t]
        \centering
        \includegraphics[width=\linewidth]{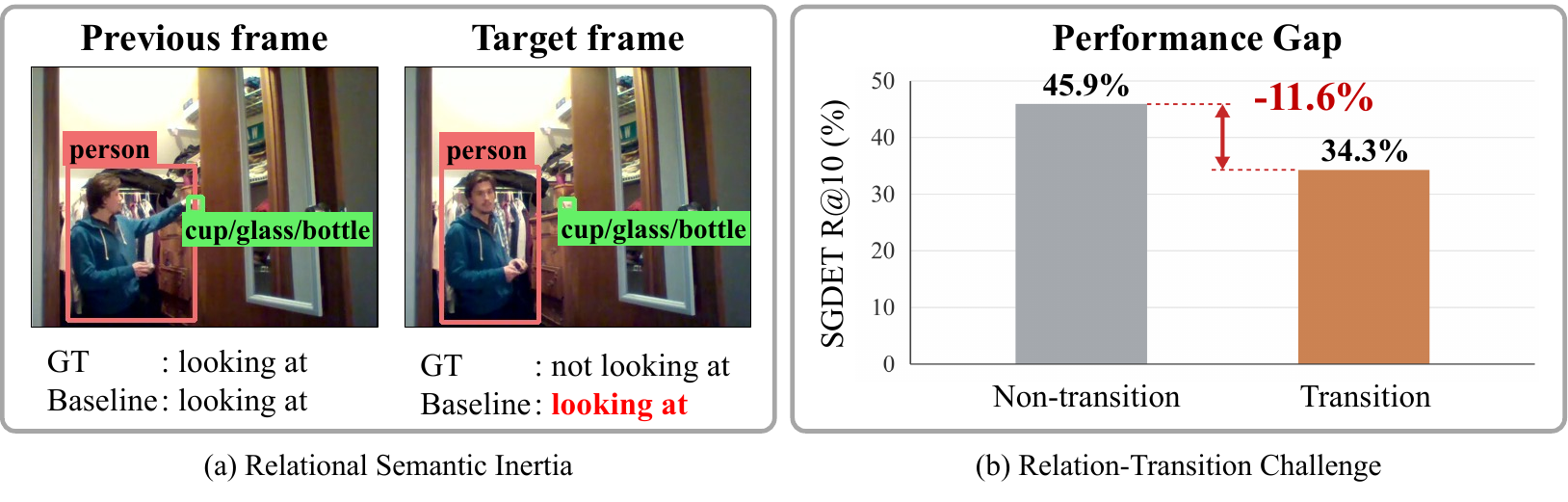}
        \caption{
            (a) Relational semantic inertia: the `person--cup/glass/bottle' pair persists across frames, but its relation changes from \emph{looking at} to \emph{not looking at}, while the baseline~\cite{wang2024oed} retains the previous predicate in the target frame.
            (b) Our reimplementation of the baseline exhibits an 11.6-point gap in SGDET R@10 between the non-transition and relation-transition frame subsets.
        }
        \label{fig:teaser}
    \end{figure}

    Despite these advances, capturing abrupt relation changes remains challenging when the same entities persist across frames.
    While existing approaches aggregate information across frames to improve predictions~\cite{cong2021sttran, feng2023dsgdetr, wang2024oed, wang2025arn, zhang2024tpt}, explicitly modeling relation changes for persistent entity pairs has received comparatively less attention.
    In such cases, a model may retain its previous prediction even after the underlying relation has changed.
    Figure~\ref{fig:teaser}(a) illustrates this behavior: the same \emph{subject--object} pair remains visible, but its relation changes from \emph{looking at} to \emph{not looking at}, while a baseline model~\cite{wang2024oed} continues to predict \emph{looking at}.
    Our reimplementation of the baseline~\cite{wang2024oed} achieves an R@10 of 34.3 on the relation-transition frame subset, compared with 45.9 on the non-transition frame subset under the with constraint setting (Fig.~\ref{fig:teaser}(b)).
    We refer to this failure mode as \emph{relational semantic inertia}.

    These observations motivate temporal modeling that preserves stable entity information while explicitly accounting for changes in relationships.
    To this end, we propose \textbf{Relation-Transition-Aware Transformer (RTAformer)}, a one-stage VidSGG framework that uses differences between adjacent-frame relation queries as explicit cues for predicate updates.
    RTAformer refines instance and relation queries in separate temporal streams to aggregate entity information and incorporate relation changes into predicate prediction.
    In the relation stream, adjacent-frame query differences are calibrated and injected before temporal aggregation.
    These differences are more informative when the queries remain grounded to consistent entity pairs across frames.
    We therefore regularize adjacent-frame instance-pair queries during training to encourage consistent grounding and help reduce representational drift in the transition cues.
    Our main contributions are as follows:
    \begin{enumerate}
        \item We characterize \emph{relational semantic inertia}, a failure mode in VidSGG where models retain previous predicate predictions despite relation changes between persistent entities.

        \item We propose RTAformer, a one-stage framework that separates entity and relation temporal refinement and uses calibrated relation-query differences as explicit cues for predicate updates, supported by cross-frame instance alignment that encourages consistent query grounding.

        \item Experiments on Action Genome~\cite{ji2020action} demonstrate improved scene graph generation performance over state-of-the-art methods, with a backbone-controlled comparison against OED~\cite{wang2024oed} showing larger gains on the relation-transition frame subset than on the non-transition frame subset.
    \end{enumerate}

\section{Related Work}

\noindent\textbf{Video Scene Graph Generation.}
    Video scene graph generation (VidSGG) extends image-based scene graph generation~\cite{yang2018graphrcnn, zellers2018motifs, tang2019vctree, cong2023reltr, li2024sgtrplus, im2024egtr} by modeling objects and their relationships across video frames.
    Earlier methods typically separate object detection from subsequent temporal and relational reasoning~\cite{teng2021trace, wang2022tpi, cong2021sttran}, whereas recent one-stage frameworks integrate instance and relation prediction~\cite{wang2024oed, wang2025arn, le2026uno}.
    OED~\cite{wang2024oed} and ARN~\cite{wang2025arn} use learnable queries to predict instances and relations from visual features, while UNO~\cite{le2026uno} adopts an object-centric slot-attention framework.
    Our work builds on the one-stage formulation by introducing explicit relation-change cues into temporal refinement.

\noindent\textbf{Temporal Aggregation and Relation Transitions.}
    Temporal aggregation uses cross-frame context to refine instance and relation predictions.
    STTran~\cite{cong2021sttran} captures inter-frame dependencies with a temporal transformer, APT~\cite{li2022apt} introduces anticipatory learning objectives, and DSG-DETR~\cite{feng2023dsgdetr} uses object tracklets to exploit long-term dependencies.
    Within one-stage frameworks, OED~\cite{wang2024oed} aggregates temporal information through cross-attention over instance and relation queries, while ARN~\cite{wang2025arn} combines temporal feature modeling with entity-predicate association reasoning.
    Complementary efforts address prediction reliability, including memory-guided debiasing in TEMPURA~\cite{nag2023tempura} and selective spatio-temporal denoising in TD$^2$-Net~\cite{lin2024td2}.
    
    More directly related to our motivation, TR$^2$~\cite{wang2023tr2} addresses time-varying relations through text-guided supervision of inter-frame relation-feature differences.
    It also uses a short-term message token to modulate the contribution of preceding-frame features during relation prediction.
    RTAformer shares the motivation of capturing relation changes but uses adjacent-frame differences as a direct input to relation-query refinement.
    Specifically, it calibrates and injects these differences before temporal aggregation within a one-stage framework, while separate instance and relation streams accommodate entity persistence and relation changes.

\noindent\textbf{Query-Based Temporal Modeling.}
    DETR~\cite{carion2020detr} formulates object detection as set prediction using learnable object queries, while subsequent methods incorporate explicit positional priors into query design~\cite{liu2022dabdetr, zhang2023dino}.
    For video understanding, queries also provide a means of incorporating temporal information.
    FAQ~\cite{cui2023faq} aggregates neighboring-frame evidence into queries for video object detection, while TrackFormer~\cite{meinhardt2022trackformer} and MOTR~\cite{zeng2022motr} propagate query states to maintain object tracks.
    In VidSGG, TPT~\cite{zhang2024tpt} uses temporal propagation to connect predictions across frames.
    RTAformer focuses on differences between adjacent-frame relation queries as cues for predicate updates.
    Because these differences can also reflect inconsistent entity-pair grounding, training-time instance alignment encourages cross-frame consistency to help reduce representational drift in the transition cues.

\section{Method}
    
    \subsection{Problem Formulation} 
    \label{sec:problem_formulation}
    Given a video sequence $\mathcal{V}=\{I_t\}_{t=1}^{T}$ consisting of $T$ temporally ordered frames, video scene graph generation (VidSGG) aims to construct a dynamic spatiotemporal representation $\mathcal{G} = \{\mathcal{G}_t\}_{t=1}^T$ that captures both the presence of physical entities and their evolving interactions. 
    Formally, each frame-wise graph is defined as 
    $\mathcal{G}_t = (\mathcal{O}_t, \mathcal{R}_t)$, where $\mathcal{O}_t = \{o_{t,i}\}_{i=1}^{N_t}$ 
    is the set of detected objects, each characterized by a spatial bounding box $b_{t,i} \in \mathbb{R}^4$ and a categorical label $c_{t,i} \in \mathcal{Y}_{obj}$, where $\mathcal{Y}_{obj}$ denotes the set of object categories.
    The relation set $\mathcal{R}_t$ comprises interaction triplets $\langle o_s, p, o_o \rangle$, where $o_s, o_o \in \mathcal{O}_t$ are the \emph{subject} and \emph{object}, and $p \in \{0,1\}^{|\mathcal{C}_{pred}|}$ is a multi-label predicate vector, where $\mathcal{C}_{pred}$ denotes the set of predicate categories.
    The objective is to learn a mapping $\mathcal{V} \mapsto \mathcal{G}$ that maximizes the posterior $P(\mathcal{G} \mid \mathcal{V})$.

    \begin{figure}[t]  
        \centering
        \includegraphics[width=\linewidth]{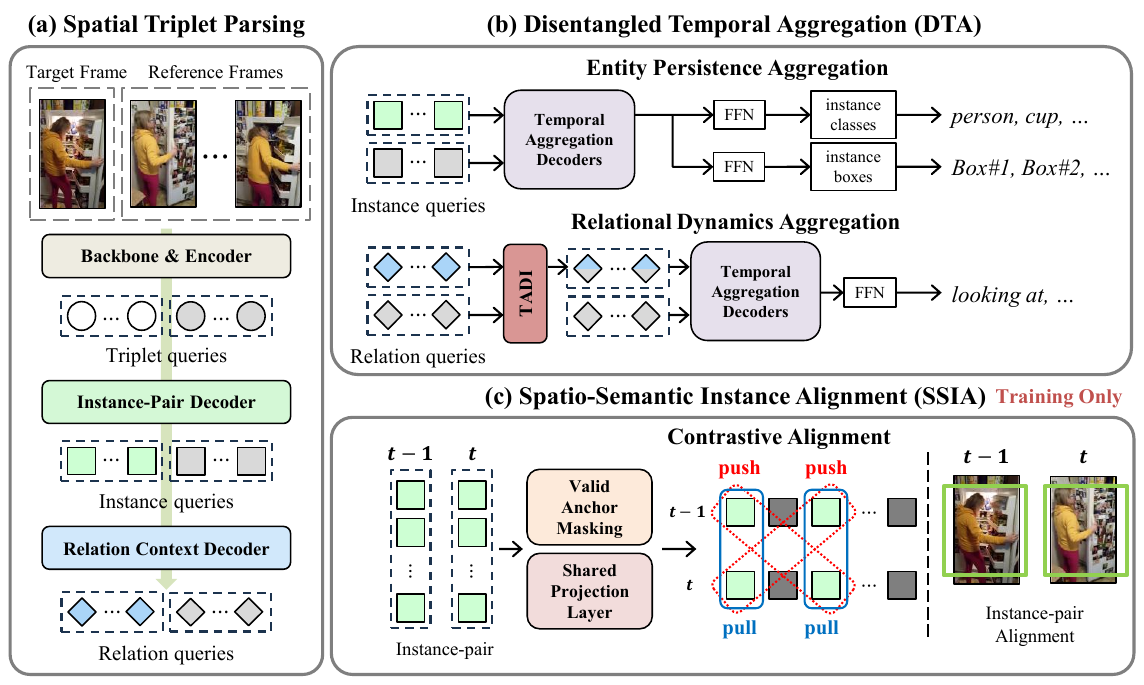}
        \caption{
            Overall framework of RTAformer.
            (a) Frame-wise spatial triplet parsing produces instance-pair and relation queries for target and reference frames.
            (b) Disentangled Temporal Aggregation (DTA) refines the target queries with separate instance and relation streams, where TADI injects relation-transition cues.
            FFN heads produce the instance and relation predictions that form the output triplets.
            (c) Spatio-Semantic Instance Alignment (SSIA) aligns adjacent-frame instance-pair queries during training.
            }
        \label{fig:framework}
    \end{figure}

    \subsection{Framework Overview} 
    \label{sec:overview} 
        As illustrated in Fig.~\ref{fig:framework}, RTAformer is a one-stage set-prediction framework that models the distinct temporal behaviors of entities and predicates in VidSGG.
        The framework consists of three key components.
        
        \noindent\textbf{(1) Frame-Wise Spatial Triplet Parsing.} 
        Given a target frame $I_{tar}$ and $N_{ref}$ reference frames $\{I^{1}_{ref}, \ldots, I^{N_{ref}}_{ref}\}$, a vision backbone~\cite{dosovitskiy2020vit, oquab2023dinov2} followed by a Transformer encoder~\cite{vaswani2017attention} produces encoded visual features for each frame. 
        These features are processed by two cascaded decoders, the \textit{Instance-Pair Decoder} and the \textit{Relation Context Decoder}, which transform $N_q$ learnable triplet queries $Q$ into frame-wise instance-pair queries $Q^{ins}$ and relation queries $Q^{rel}$.
        
        \noindent\textbf{(2) Disentangled Temporal Aggregation (DTA).} 
        To mitigate relational semantic inertia, DTA refines the target queries $Q^{ins}_{tar}$ and $Q^{rel}_{tar}$ in separate instance and relation streams.
        The instance stream aggregates entity information across frames, while the
        relation stream applies \textit{Transition-Aware Delta Injection (TADI)} to inject inter-frame relation-change cues into predicate decoding.
        
        \noindent\textbf{(3) Spatio-Semantic Instance Alignment (SSIA).} 
        During training, SSIA applies contrastive regularization to adjacent-frame instance-pair queries to improve cross-frame query consistency and reduce representational drift in the displacements used by TADI.

    \subsection{Frame-Wise Spatial Triplet Parsing}
    \label{sec:spatial_parsing}
        Following the DETR-like set prediction paradigm~\cite{carion2020detr}, we associate $N_q$ learnable triplet queries $Q \in \mathbb{R}^{N_q \times D}$ with potential \emph{subject--object} interactions, where $D$ denotes the hidden dimension. 
        Unlike standard object detection queries that represent individual entities, each triplet query represents a candidate \emph{subject--object} interaction.
        The cascaded decoders first produce instance-pair representations and then condition relation decoding on them.        
        For notational simplicity, we describe the parsing process for a generic frame $I_t$. 
        The same spatial parser applies identically whether $t$ denotes the target frame ($t{=}tar$) or a reference frame.

        For each frame $I_t$, a vision backbone (\eg, ViT~\cite{dosovitskiy2020vit, oquab2023dinov2}) and Transformer encoder~\cite{vaswani2017attention} produce encoded visual features $F_t$. 
        The \textit{Instance-Pair Decoder} ($\mathrm{Decoder}_{ins}$) first transforms the initial queries $Q$ into instance-pair representations by aggregating spatial context from $F_t$.
        Since relational semantics are inherently conditioned on the identity and spatial configuration of the underlying \emph{subject--object} pair, the \textit{Relation Context Decoder} ($\mathrm{Decoder}_{rel}$) subsequently takes $Q_t^{ins}$ as its query input to produce relation-aware representations: 
        \begin{equation}
        \begin{aligned}
            Q_t^{ins} &= \mathrm{Decoder}_{ins}(Q,\; F_t), \\
            Q_t^{rel} &= \mathrm{Decoder}_{rel}(Q_t^{ins},\; F_t).
        \end{aligned}
        \label{eq:spatial_parsing}
        \end{equation}
        Each decoder layer follows the standard DETR decoder design~\cite{carion2020detr}, using self-attention to exchange contextual information among triplet queries and cross-attention to retrieve spatial evidence from $F_t$~\cite{liu2022dabdetr,meng2021conddetr,carion2020detr}, followed by a feed-forward network.

        The resulting $Q_t^{ins}$ and $Q_t^{rel}$ encode frame-level instance and relation contexts, respectively, and serve as inputs to the Disentangled Temporal Aggregation module described in Sec.~\ref{sec:dta}.

    \subsection{Disentangled Temporal Aggregation}
    \label{sec:dta}
        Entity identities tend to persist across frames, whereas predicates can change abruptly for the same \emph{subject--object} pair.
        \textit{Disentangled Temporal Aggregation (DTA)} therefore processes instance and relation queries in separate temporal streams.
        The instance stream aggregates persistent entity evidence, while the relation stream incorporates relation-transition cues before temporal aggregation.

        Given $N_{ref}$ reference frames, we apply the spatial parser to each and concatenate the resulting instance and relation queries into reference query sets
        $Q_{ref}^{ins},\,Q_{ref}^{rel} \in \mathbb{R}^{N_{ref}N_q \times D}$,
        where the superscript denotes the stream.
        The preceding frame in the input sequence, indexed by $tar{-}1$ and included in the reference set, provides $Q_{prev}^{rel} := Q_{tar-1}^{rel}$ and $Q_{prev}^{ins} := Q_{tar-1}^{ins}$ for computing the relation-query displacement and confidence gate in TADI, respectively.
        
        Each stream contains $N_d{=}3$ Temporal Aggregation Decoders.
        At stage $d$, the decoder applies self-attention to the target queries, cross-attention to the selected reference queries, and a feed-forward network~\cite{vaswani2017attention}.
        The two streams use independently scored reference query sets:
        \begin{equation}
            Q_{tar}^{u,(d)}
            =
            \mathrm{Decoder}^{u,(d)}_{temp}\!\left(
                Q_{tar}^{u,(d-1)},\;
                Q_{ref}^{u,(d)}
            \right),
            \quad u\in\{ins,\,rel\},
            \label{eq:temporal_agg}
        \end{equation}
        where $Q_{ref}^{u,(d)}=\mathrm{TopK}(Q_{ref}^{u},K_d)$ selects the top-$K_d$ reference queries according to the stream-specific scores defined below.
        We set $K_d=\alpha_d N_{ref}N_q$ with $(\alpha_1,\alpha_2,\alpha_3)=(0.8,0.5,0.3)$, progressively retaining the highest-scoring temporal candidates.
        The instance stream is initialized with $Q_{tar}^{ins,(0)}=Q_{tar}^{ins}$, while the relation stream is initialized with $Q_{tar}^{rel,(0)}=\widehat{Q}_{tar}^{rel}$, where $\widehat{Q}_{tar}^{rel}$ denotes the TADI-injected queries defined in Eq.~\eqref{eq:tadi}.

        \noindent\textbf{Instance stream.}\;
        The instance stream prioritizes high-confidence reference queries to aggregate persistent entity information.
        Each of the $N_{ref}N_q$ pooled reference instance queries, indexed by $j$, is scored via the object classification head $\phi_{obj}$ as:
        \begin{equation}
            s_j^{ins} = \log(P_j^{obj}+\epsilon),
            \quad
            P_j^{obj} =
            \max_{c\in\mathcal{Y}_{obj}}
            \mathrm{softmax}
            (\phi_{obj}(Q_{ref,j}^{ins}))_c,
        \label{eq:ins_score}
        \end{equation}
        where $\epsilon$ is a small constant for numerical stability.
        The top-$K_d$ candidates form $Q_{ref}^{ins,(d)}$.
        The final instance queries $Q_{tar}^{ins,*} = Q_{tar}^{ins,(N_d)}$ are obtained via Eq.~\eqref{eq:temporal_agg}.

        \begin{figure}[t]
            \centering
            \includegraphics[width=1.0\linewidth]{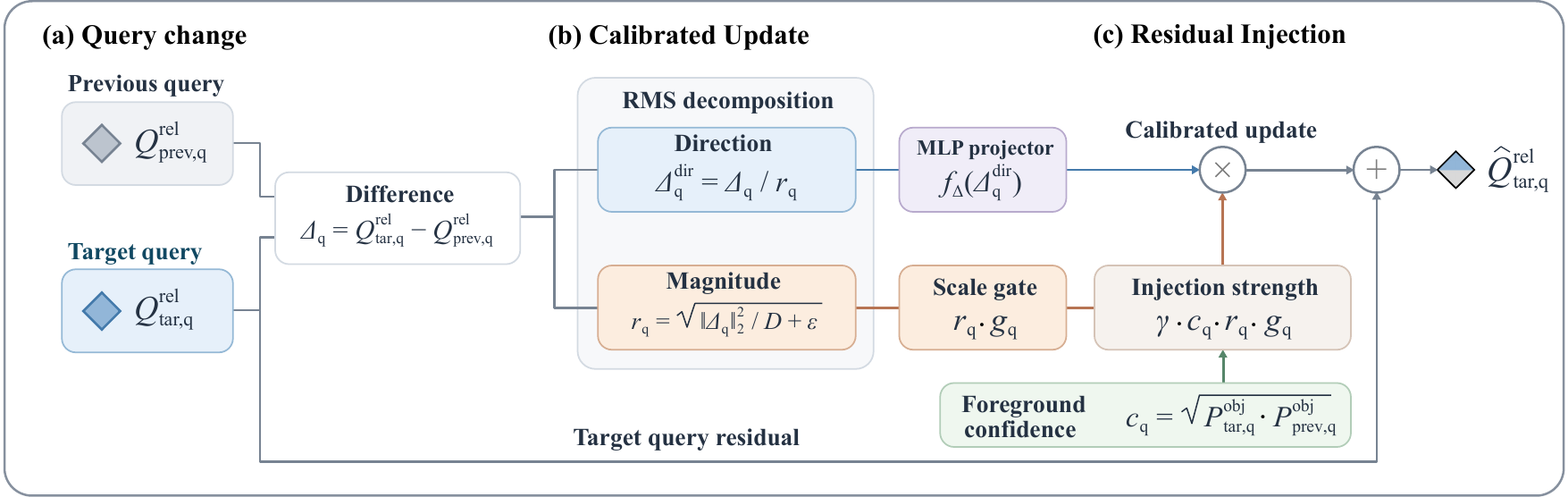}
            \caption{
            Transition-Aware Delta Injection (TADI).
            TADI uses the adjacent-frame relation-query displacement
            $\Updelta_q = Q^{rel}_{tar,q} - Q^{rel}_{prev,q}$ as a transition cue.
            The displacement is decomposed into its numerically stabilized root-mean-square (RMS) magnitude $r_q$ and RMS-normalized direction $\Updelta_q^{dir}$, while the scale gate $g_q$ and foreground confidence $c_q$ modulate the injection strength.
            }
            \label{fig:tadi}
        \end{figure}

        \noindent\textbf{Relation stream with TADI.}\;
        The relation stream applies \textit{Transition-Aware Delta Injection (TADI)} to the target relation queries before temporal aggregation.
        TADI uses differences between same-index relation queries across adjacent frames as candidate cues for predicate updates.
        As illustrated in Fig.~\ref{fig:tadi}, for each query index $q$, the displacement is computed as:
        \begin{equation}
            \Updelta_q = Q_{tar,q}^{rel} - Q_{prev,q}^{rel},
        \label{eq:delta}
        \end{equation}
        where $Q_{tar}^{rel}$ denotes the target-frame relation queries from Eq.~\eqref{eq:spatial_parsing}.
        To support these same-index comparisons, SSIA encourages cross-frame consistency of the underlying instance-pair representations during spatial training (Sec.~\ref{sec:ssia}).
        Rather than injecting raw $\Updelta_q$, TADI separates its magnitude and direction and computes a magnitude-dependent scale gate:
        \begin{equation}
        \begin{gathered}
            r_q = \sqrt{\tfrac{1}{D}\|\Updelta_q\|_2^2+\epsilon},\\
            \Updelta_q^{dir} = \Updelta_q / r_q,\\
            g_q = 1+\tanh(f_g(\log(1+r_q))),
        \end{gathered}
        \label{eq:decompose}
        \end{equation}
        where $r_q$ is the numerically stabilized root-mean-square (RMS) magnitude of $\Updelta_q$, $\Updelta_q^{dir}$ is the corresponding RMS-normalized direction, and $g_q$ is a bounded scale gate predicted by a two-layer MLP $f_g$.
        
        We modulate the update using foreground object confidence from both frames, reducing its contribution when either query has low confidence:
        \begin{equation}
            c_q = \sqrt{P_{tar,q}^{obj} \cdot P_{prev,q}^{obj}},
        \label{eq:conf_gate}
        \end{equation}
        where $P_{tar,q}^{obj}$ and $P_{prev,q}^{obj}$ are computed analogously to Eq.~\eqref{eq:ins_score}. The transition-aware target relation query is then:
        \begin{equation}
            \widehat{Q}_{tar,q}^{rel}
            = Q_{tar,q}^{rel}
            + \gamma\,c_q\,r_q\,g_q\,
              f_\Delta(\Updelta_q^{dir}),
        \label{eq:tadi}
        \end{equation}
        where $f_\Delta$ is a learnable MLP projector and 
        $\gamma = \gamma_{\max}\sigma(\rho_\gamma)$ is a 
        learnable global injection strength bounded by 
        $\gamma_{\max}$.
        Here, $f_\Delta(\Updelta_q^{dir})$ maps the normalized displacement direction to an update vector, while the scalar $c_q r_q g_q$ scales this update according to foreground confidence and displacement magnitude.  
        The resulting set of TADI-injected queries is
$\widehat{Q}_{tar}^{rel} = \{\widehat{Q}_{tar,q}^{rel}\}_{q=1}^{N_q}$.

        Reference relation queries are ranked using their object and predicate confidence, together with a displacement-saliency term computed from the target and preceding frames:
        \begin{equation}
            s_j^{rel}
            =
            \log\!\left(\max(P_j^{obj},\epsilon)\right)
            +
            \log\!\left(\max(P_j^{rel},\epsilon)\right)
            +
            \log\!\left(1+\eta\,\bar{a}_{q(j)}\right),
        \label{eq:rel_score}
        \end{equation}
        where $P_j^{rel}$ is the product of the maximum predicate probabilities from the attention, spatial, and contacting families, and $q(j)$ maps reference candidate $j$ to its query index.
        The saliency $\bar{a}_{q(j)}$ is the clipped, query-mean-normalized form of $P_{tar,q(j)}^{obj}r_{q(j)}g_{q(j)}$, and $\eta$ controls its contribution.
        The top-$K_d$ candidates form $Q_{ref}^{rel,(d)}$, 
        and the final relation queries $Q_{tar}^{rel,*} = Q_{tar}^{rel,(N_d)}$ are obtained via Eq.~\eqref{eq:temporal_agg}.
        The refined $Q_{tar}^{ins,*}$ and $Q_{tar}^{rel,*}$ are passed to temporal prediction heads for object classification, box regression, and predicate classification.

    \subsection{Spatio-Semantic Instance Alignment}
    \label{sec:ssia}
        Because TADI computes displacement between same-index queries across adjacent frames, cross-frame query consistency is important for reducing representational drift in the displacement signal.
        Following prior work on same-index contrastive alignment for temporal representation learning~\cite{manasyan2025slotcontrast}, \textit{Spatio-Semantic Instance Alignment (SSIA)} applies contrastive regularization to adjacent-frame instance-pair queries during spatial training.
        This encourages cross-frame consistency of instance-pair queries and introduces no inference overhead.

    \noindent\textbf{Anchor construction.}\;
    Bipartite matching between the predicted triplet queries and ground-truth triplets at frame $t$ yields $M$ matched query indices $\{q_m\}_{m=1}^M$.
    Here, $c_{t,m}$ and $b_{t,m}$ denote the category and bounding box of the object entity in the ground-truth triplet matched to query $q_m$ (Sec.~\ref{sec:problem_formulation}).
    For each $q_m$, we pair $Q_{t,q_m}^{ins}$ with the same-index embedding $Q_{t-1,q_m}^{ins}$ from frame $t{-}1$ as a candidate positive, without a separate matching step.
    We use ground-truth object persistence to select anchors for same-index contrastive alignment.
    Specifically, an anchor is retained when a ground-truth object $j$ in frame $t{-}1$ shares the category of the object in the matched triplet and sufficiently overlaps its box:
    \begin{equation}
        \mathcal{M}_m = \ind\!\left(
            \max_{j:\, c_{t-1,j} = c_{t,m}}
            \mathrm{IoU}(b_{t,m},\, b_{t-1,j}) > \tau
        \right).
    \label{eq:validity}
    \end{equation}
    We set $\mathcal{M}_m=0$ when no object of category $c_{t,m}$ exists in frame $t{-}1$.
    
    \noindent\textbf{Contrastive objective.}\;
    A two-layer MLP projection head $f_{proj}$ maps instance-pair queries 
    $Q_{t,q_m}^{ins}$, $Q_{t-1,q_m}^{ins}$ to $\ell_2$-normalized embeddings 
    \mbox{$z_{t,m}, z_{t-1,m} \in \mathbb{R}^D$}.
    Let $\mathcal{A} = \{m \mid \mathcal{M}_m = 1\}$ be the valid anchor set.
    The concatenated embedding bank of size $2M$ is
    $\mathcal{Z} = \{z_{t,1}, \ldots, z_{t,M},
    z_{t-1,1}, \ldots, z_{t-1,M}\}$, indexed
    such that $z_m = z_{t,m}$ for $m \leq M$ and
    $z_{m+M} = z_{t-1,m}$ for $m \leq M$.
    The validity mask restricts the anchor set, while all $2M$ embeddings remain in the contrastive bank.
    SSIA applies a symmetric InfoNCE loss~\cite{oord2018cpc} over valid 
    anchors:
    \begin{equation}
        \mathcal{L}_{SSIA}
        = -\frac{1}{2|\mathcal{A}|}\sum_{m\in\mathcal{A}}
        \left[
        \ell(z_{t,m},\,z_{t-1,m},\,m)
        +
        \ell(z_{t-1,m},\,z_{t,m},\,m{+}M)
        \right],
    \label{eq:ssia}
    \end{equation}
    where 
    $\ell(a,b,i) = \log\frac{\exp(a^\top b / \kappa)}
    {\sum_{k \neq i} \exp(a^\top z_k / \kappa)}$
    and $\kappa$ is the temperature.
    If $\mathcal{A}=\varnothing$, we set $\mathcal{L}_{SSIA}=0$.

    \subsection{Optimization and Inference}
    \label{sec:optimization}
    
    \noindent\textbf{Set-prediction objective.}\;
    We assign triplet queries to ground-truth annotations using Hungarian
    matching~\cite{carion2020detr, wang2024oed}, with a weighted cost combining
    object and predicate classification with $\ell_1$ and GIoU terms for
    subject and object boxes.
    The training loss is
    \begin{equation}
        \mathcal{L}_{SGG}
        = \lambda_{obj}\mathcal{L}_{obj}
        + \lambda_{rel}\mathcal{L}_{rel}
        + \lambda_{L1}\mathcal{L}_{L1}^{box}
        + \lambda_{giou}\mathcal{L}_{giou}^{box},
    \end{equation}
    where $\lambda$ denotes the loss weights.
    Object classification supervises all queries, with unmatched queries
    assigned to the no-object class.
    Predicate and box losses apply only to matched queries.
    The box losses supervise both subject and object boxes, while
    $\mathcal{L}_{rel}$ sums the classification losses for the attention,
    spatial, and contacting families.

    \noindent\textbf{Decoupled optimization and one-stage inference.}\;
    Following OED~\cite{wang2024oed}, we train RTAformer in two phases, first learning the spatial representations and then optimizing temporal refinement.
    In Phase~1, the spatial parser is trained with the backbone frozen, using frame-level supervision and SSIA regularization:
    \begin{equation}
    \mathcal{L}_{stage1}
    =
    \mathcal{L}^{spatial}_{SGG}
    +
    \lambda_{ssia}\mathcal{L}_{SSIA},
    \end{equation}
    where $\mathcal{L}^{spatial}_{SGG}$ denotes $\mathcal{L}_{SGG}$ computed from the frame-wise spatial parsing outputs.
    In Phase~2, the backbone and spatial modules are frozen, and only the DTA module and temporal prediction heads are optimized on sequence-level clips:
    \begin{equation}
        \mathcal{L}_{stage2}
        =
        \mathcal{L}^{temporal}_{SGG},
    \end{equation}
    where $\mathcal{L}^{temporal}_{SGG}$ denotes 
    $\mathcal{L}_{SGG}$ computed from the temporally 
    refined DTA outputs.
    This two-phase schedule is used only during training.
    At inference, RTAformer performs spatial triplet parsing, DTA, and final prediction without external trackers or temporal post-processing.
\section{Experiments}

    \noindent\textbf{Dataset.}
    Following prior VidSGG work~\cite{cong2021sttran, li2022apt, nag2023tempura, lin2024td2, wang2024oed, wang2025arn, zhuang2025stabile}, we evaluate our method on Action Genome (AG)~\cite{ji2020action}, the standard benchmark for video scene graph generation.
    AG contains 234,253 annotated frames sampled from 9,848 videos, with 476,229 object bounding boxes and 1,715,568 relation annotations across 35 object categories excluding \emph{person} and 25 predicate categories.
    The predicates are grouped into three semantic types: \emph{attention}, \emph{spatial}, and \emph{contacting}.

    \noindent\textbf{Evaluation Metrics.}
    \label{sec:metrics}
    We evaluate our model under the standard scene graph detection (SGDET) protocol.
    SGDET requires the model to jointly localize object instances, classify their object categories, and predict the predicates between \emph{subject--object} pairs from video frames.
    We report recall@K (R@K) with $K \in \{10,20,50\}$ under two graph construction settings: \textbf{With Constraint} and \textbf{No Constraint}.
    The With Constraint setting restricts each \emph{subject--object} pair to its highest-scoring predicate prediction, whereas the No Constraint setting allows multiple predicate predictions for the same pair.
    We additionally report mean recall@K (mR@K), which averages recall across predicate categories to give each category equal weight.
    
    \noindent\textbf{Implementation Details.}
    Our framework follows recent query-based one-stage VidSGG architectures~\cite{wang2024oed,wang2025arn}.
    Following UNO~\cite{le2026uno}, we use the same frozen DINOv2 vision transformer (ViT) backbone~\cite{dosovitskiy2020vit,oquab2023dinov2}.
    The transformer encoder, instance-pair decoder, and relation decoder each contain six transformer layers, and the model uses $N_q = 100$ learnable triplet queries.
    In DTA, the instance and relation streams each contain three temporal aggregation decoders, each implemented as a single transformer decoder layer.
    For SSIA, we set $\kappa = 0.1$, $\tau = 0.3$, and $\lambda_{\mathrm{SSIA}} = 0.1$.
    We train the model using the AdamW optimizer on four NVIDIA RTX A6000 GPUs.
    Unless otherwise specified, ablation studies use ViT-B/14, while the main comparisons report ViT-S/14, ViT-B/14, and ViT-L/14 variants.
    Additional implementation details are provided in the supplementary material.
    
    \subsection{Comparison with State-of-the-Art Methods}
    
    \begin{table}[t]
        \caption{
            Experimental results on Action Genome under the SGDET protocol.
            We report Recall@K (R@K) and mean Recall@K (mR@K) under With Constraint and No Constraint settings. 
            `--' indicates that the metric is not reported.
        }
        \label{tab:sota_main}
        \centering
        \scriptsize
        \setlength{\tabcolsep}{2.0pt}
        \renewcommand{\arraystretch}{1.05}
        \resizebox{\textwidth}{!}{
        \begin{tabular}{@{}l c ccc | ccc | ccc | ccc@{}}
        \toprule
        \multirow{2}{*}[-0.5ex]{\textbf{Method}}
        & \multirow{2}{*}[-0.5ex]{\textbf{Backbone}}
        & \multicolumn{6}{c}{\textbf{With Constraint}}
        & \multicolumn{6}{c}{\textbf{No Constraint}} \\
        
        \cmidrule(l{0.1pt}r{0.1pt}){3-8}
        \cmidrule(l{0.1pt}r{0.1pt}){9-14}
        
        &
        & \textbf{R@10} & \textbf{R@20} & \textbf{R@50}
        & \textbf{mR@10} & \textbf{mR@20} & \textbf{mR@50}
        & \textbf{R@10} & \textbf{R@20} & \textbf{R@50}
        & \textbf{mR@10} & \textbf{mR@20} & \textbf{mR@50} \\
        
        \midrule

        \rowcolor{multistagebg}
        \multicolumn{14}{c}{\textbf{Multi-stage methods}} \\
        \midrule
        STTran~\cite{cong2021sttran} & ResNet-101
        & 25.2 & 34.1 & 37.0
        & 16.6 & 20.8 & 22.2 
        & 24.6 & 36.2 & 48.8
        & 20.9 & 29.7 & 39.2 \\

        APT~\cite{li2022apt} & ResNet-101
        & 26.3 & 36.1 & 38.3
        & -- & -- & --
        & 25.7 & 37.9 & 50.1
        & -- & -- & -- \\

        TR$^2$~\cite{wang2023tr2} & ResNet-101
        & 26.8 & 35.5 & 38.3
        & -- & -- & --
        & 27.8 & 39.2 & 50.0
        & -- & -- & -- \\

        TEMPURA~\cite{nag2023tempura} & ResNet-101
        & 28.1 & 33.4 & 34.9
        & 18.5 & 22.6 & 23.7
        & 29.8 & 38.1 & 46.4
        & 24.7 & 33.9 & 43.7 \\

        DSG-DETR~\cite{feng2023dsgdetr} & ResNet-101
        & 30.3 & 34.8 & 36.1
        & -- & -- & --
        & 32.1 & 40.9 & 48.3
        & -- & -- & -- \\

        TD$^2$-Net~\cite{lin2024td2} & ResNet-101
        & 28.7 & -- & 37.1
        & 20.4 & -- & 26.1
        & 30.5 & -- & 49.3
        & 27.9 & -- & 46.3 \\

        STABILE~\cite{zhuang2025stabile} & ResNet-101
        & 29.9 & 34.2 & 35.0
        & 22.8 & 25.6 & 26.5
        & 33.7 & 41.2 & 47.7
        & 29.1 & 36.5 & 43.9 \\

        DIFFVSGG~\cite{chen2025diffvsgg} & ResNet-50+DETR
        & 34.7 & 41.9 & 47.3
        & 24.7 & 27.3 & 28.4
        & 37.4 & 45.1 & 53.1
        & 29.7 & 37.9 & 46.4 \\
        \midrule

        \rowcolor{multistagebg}
        \multicolumn{14}{c}{\textbf{One-stage methods}} \\
        \midrule
        OED~\cite{wang2024oed} & ResNet-50
        & 33.5 & 40.9 & 48.9
        & 21.5 & 27.4 & 33.4
        & 35.3 & 44.0 & 51.8
        & 27.2 & 40.5 & 50.4 \\

        ARN~\cite{wang2025arn} & ResNet-50
        & 35.1 & 41.8 & 47.2
        & 23.2 & 29.0 & 33.6
        & 37.6 & 46.8 & 54.1
        & 29.7 & 41.9 & 51.9 \\

        UNO~\cite{le2026uno} & ResNet-50
        & 35.4 & 42.2 & 49.5
        & -- & -- & --
        & 36.6 & 46.1 & 53.9
        & -- & -- & -- \\

        UNO~\cite{le2026uno} & ViT-S/14
        & 36.7 & 43.1 & 50.2
        & -- & -- & --
        & 37.5 & 47.5 & 54.5
        & -- & -- & -- \\

        UNO~\cite{le2026uno} & ViT-B/14
        & 38.2 & 44.7 & 51.9
        & -- & -- & --
        & 39.9 & 48.2 & 56.3
        & -- & -- & -- \\

        UNO~\cite{le2026uno} & ViT-L/14
        & 39.3 & 45.2 & 53.8
        & -- & -- & --
        & 40.8 & 49.7 & 57.1
        & -- & -- & -- \\
        \midrule

        \textbf{Ours} & ViT-S/14
        & 37.7 & 45.1 & 52.3
        & 24.6 & 30.9 & 36.2
        & 40.0 & 49.7 & 57.5
        & 30.5 & 45.0 & 56.0 \\

        \textbf{Ours} & ViT-B/14
        & 42.2 & 50.1 & 57.6
        & 28.8 & 35.6 & 41.3
        & 44.8 & 55.0 & 63.2
        & 35.3 & 50.7 & 61.9 \\

        \textbf{Ours} & ViT-L/14
        & \textbf{44.6} & \textbf{52.9} & \textbf{60.2}
        & \textbf{30.5} & \textbf{38.3} & \textbf{44.2}
        & \textbf{47.2} & \textbf{57.8} & \textbf{65.9}
        & \textbf{37.2} & \textbf{53.5} & \textbf{64.4} \\
        \bottomrule
        \end{tabular}
        }
        \vspace{3pt}
    \end{table}

    We compare RTAformer with recent VidSGG methods, including both multi-stage pipelines and one-stage frameworks.
    Table~\ref{tab:sota_main} reports SGDET R@K and class-balanced mR@K under both \textit{With Constraint} and \textit{No Constraint} settings. 
    For mR@K, methods without reported values~\cite{li2022apt, wang2023tr2, feng2023dsgdetr, le2026uno} are marked as `--' and are not considered in the mR@K comparison.

    With matched frozen DINOv2 backbones~\cite{oquab2023dinov2}, RTAformer consistently outperforms UNO~\cite{le2026uno} across all backbone scales and both constraint settings.
    With ViT-L/14, RTAformer achieves R@10/20/50 of 44.6/52.9/60.2 under \textit{with constraint} and 47.2/57.8/65.9 under \textit{no constraint}, improving R@10 over UNO by 5.3 and 6.4 percentage points, respectively.
    The R@10 margin increases with backbone scale under both settings, from 1.0/2.5 percentage points with ViT-S/14 to 5.3/6.4 percentage points with ViT-L/14.
    The ViT-L/14 variant also achieves the highest R@K among all methods reported in Table~\ref{tab:sota_main}.

    We further compare mR@K among methods that report it.
    With ViT-L/14, RTAformer attains the highest reported mR@K across all evaluated cutoffs and settings, reaching 30.5/38.3/44.2 under \textit{with constraint} and 37.2/53.5/64.4 under \textit{no constraint}, outperforming both ARN~\cite{wang2025arn} and DIFFVSGG~\cite{chen2025diffvsgg}.
    The improvements extend to class-averaged recall, complementing the gains in overall R@K.
        
    \subsection{Ablation Studies}    

        \noindent\textbf{Effect of SSIA and DTA.}\;
        Table~\ref{tab:ablation_modules} reports module-level contributions.
        Both SSIA and DTA improve over the spatial baseline, and their combination achieves the highest recall across all evaluated cutoffs and settings.
        SSIA remains beneficial when DTA is enabled, increasing R@10 from 41.2 to 42.2 under \textit{with constraint}.
        
        \noindent\textbf{Effect of Transition-Aware Delta Injection.}\; Table~\ref{tab:ablation_temporal} compares transition cue injection designs within the relation stream.
        Temporal aggregation without injection already improves over the spatial+SSIA baseline, supporting the benefit of temporal context.
        Injecting relation-query displacement further improves performance, with raw delta consistently outperforming direction-only injection.
        TADI achieves the highest recall across all evaluated cutoffs and settings, supporting the effectiveness of the complete calibrated injection design.

        \begin{table}[t]
        \centering
    
        \begin{minipage}[t]{0.485\textwidth}
            \centering
            \captionof{table}{
                Module ablation.
                Starting from the frame-wise spatial baseline, we evaluate SSIA and DTA individually and jointly.
                The spatial baseline performs frame-wise triplet parsing without temporal refinement.
                DTA denotes the full temporal module with TADI.
                All variants use the ViT-B/14 backbone.
            }
            \label{tab:ablation_modules}
    
            \vspace{2pt}
    
            \scriptsize
            \setlength{\tabcolsep}{1.6pt}
            \renewcommand{\arraystretch}{1.0}
    
            \resizebox{\linewidth}{!}{
            \begin{tabular}{c c c c c c c c c}
                \toprule
                \multirow{2}{*}[-0.5ex]{Spatial}
                & \multirow{2}{*}[-0.5ex]{SSIA}
                & \multirow{2}{*}[-0.5ex]{DTA}
                & \multicolumn{3}{c}{With Constraint}
                & \multicolumn{3}{c}{No Constraint} \\
                \cmidrule(lr){4-6}
                \cmidrule(lr){7-9}
                & & & R@10 & R@20 & R@50
                & R@10 & R@20 & R@50 \\
                \midrule
                \checkmark & -- & --
                & 39.7 & 47.5 & 53.7
                & 42.0 & 52.6 & 61.2 \\
    
                \checkmark & \checkmark & --
                & 40.8 & 48.1 & 53.9
                & 43.3 & 53.6 & 61.4 \\
    
                \checkmark & -- & \checkmark
                & 41.2 & 49.4 & 56.8
                & 43.8 & 54.3 & 62.8 \\
    
                \checkmark & \checkmark & \checkmark
                & \textbf{42.2} & \textbf{50.1} & \textbf{57.6}
                & \textbf{44.8} & \textbf{55.0} & \textbf{63.2} \\
                \bottomrule
            \end{tabular}
            }
        \end{minipage}
        \hfill
        \begin{minipage}[t]{0.485\textwidth}
            \centering
            \captionof{table}{
                Ablation on relation-stream transition injection design.
                Starting from the spatial+SSIA baseline, we compare temporal aggregation without transition injection and three relation-displacement injection variants.
                All variants use the same spatial branch and ViT-B/14 backbone.
            }
            \label{tab:ablation_temporal}
    
            \vspace{1pt}
    
            \footnotesize
            \setlength{\tabcolsep}{1.8pt}
            \renewcommand{\arraystretch}{1.0}
    
            \resizebox{\linewidth}{!}{
            \begin{tabular}{l ccc ccc}
                \toprule
                \multirow{2}{*}[-0.5ex]{Variant}
                & \multicolumn{3}{c}{With Constraint}
                & \multicolumn{3}{c}{No Constraint} \\
                \cmidrule(lr){2-4}
                \cmidrule(lr){5-7}
                & R@10 & R@20 & R@50
                & R@10 & R@20 & R@50 \\
                \midrule
    
                Spatial + SSIA
                & 40.8 & 48.1 & 53.9
                & 43.3 & 53.6 & 61.4 \\
    
                + Temporal Agg. (no inj.)
                & 41.3 & 49.0 & 56.1
                & 43.9 & 54.1 & 61.7 \\
    
                \midrule
    
                + Direction-only Inj.
                & 41.7 & 49.5 & 56.9
                & 44.2 & 54.1 & 62.0 \\
    
                + Raw Delta Inj.
                & 41.9 & 49.8 & 57.2
                & 44.4 & 54.4 & 62.4 \\
    
                + TADI
                & \textbf{42.2} & \textbf{50.1} & \textbf{57.6}
                & \textbf{44.8} & \textbf{55.0} & \textbf{63.2} \\
                \bottomrule
            \end{tabular}
            }
        \end{minipage}
    \end{table}

    \subsection{Analysis of Relation Transitions}
        We examine two questions: (i) whether relation-transition frames are particularly challenging for a baseline model, and (ii) whether adjacent-frame relation-query differences provide useful cues for these transitions.

        \noindent\textbf{Relation-transition subset construction.}            \label{sec:transition_subset_eval}
            To focus on relation changes involving persistent entities, we construct evaluation subsets from adjacent annotated frames in the AG test split.
            We retain a \emph{subject--object} pair across frames when the entities satisfy an IoU threshold of 0.3 and the object category remains consistent.
            A pair adjacency is labeled as a transition when at least one relation family (\textit{attention}, \textit{spatial}, or \textit{contacting}) has different,
            disjoint predicate sets across the two frames.
            Changes with partially overlapping predicate sets do not satisfy this criterion.
            Among 105,356 persistent pair adjacencies, 35,523 (33.7\%) satisfy the transition criterion.
            Among target frames containing at least one persistent subject--object pair, a frame is classified as relation-transition if any such pair satisfies the transition criterion, and as non-transition otherwise. This yields 25,883 relation-transition frames and 21,474 non-transition frames.

            \begin{table}[t]
            \caption{
                Controlled comparison with OED using the same frozen DINOv2 ViT-B/14 backbone.
                WC and NC denote the with constraint and no constraint settings, respectively.
                Recall entries report R@10/R@20.
            }
            \label{tab:backbone_controlled}
            \centering
            \scriptsize
            \setlength{\tabcolsep}{1.7pt}
            \renewcommand{\arraystretch}{1.05}
        
            \resizebox{\textwidth}{!}{
            \begin{tabular}{@{}l cc cc cc cc c c@{}}
                \toprule
                \multirow{2}{*}{Method}
                & \multicolumn{2}{c}{Full AG}
                & \multicolumn{2}{c}{Transition Subset}
                & \multicolumn{2}{c}{Non-transition Subset}
                & \multicolumn{2}{c}{Parameters (M)}
                & \multirow{2}{*}{GFLOPs}
                & \multirow{2}{*}{FPS} \\
                \cmidrule(lr){2-3}
                \cmidrule(lr){4-5}
                \cmidrule(lr){6-7}
                \cmidrule(lr){8-9}
        
                & WC & NC
                & WC & NC
                & WC & NC
                & Total & Temporal
                & & \\
                \midrule
        
                OED~\cite{wang2024oed}
                & 38.7/46.6 & 41.1/51.4
                & 34.3/42.7 & 37.4/48.0
                & 45.9/53.7 & 48.9/58.9
                & 126.8 & 12.9
                & 1041.6 & 31.43 \\
        
                \textbf{Ours}
                & \textbf{42.2/50.1} & \textbf{44.8/55.0}
                & \textbf{39.1/47.7} & \textbf{41.5/52.9}
                & \textbf{47.6/55.2} & \textbf{50.8/60.5}
                & 123.8 & 9.9
                & 1041.0 & 31.35 \\
        
                \bottomrule
            \end{tabular}
            }
            \vspace{4pt}
        \end{table}

        \begin{figure}[t]
            \centering
            \includegraphics[width=\linewidth]{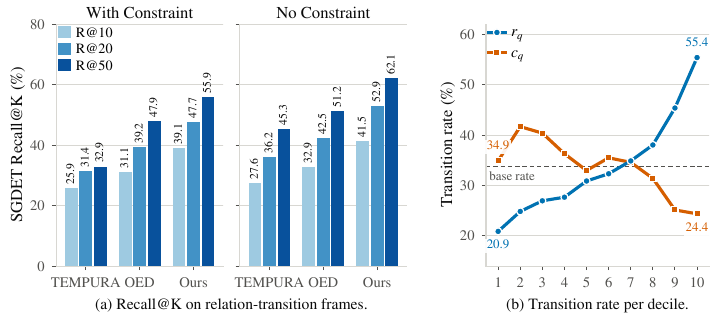}
            \caption{
            Relation-transition analysis.
            (a) Recall on the relation-transition frame subset, comparing RTAformer with ViT-B/14 against OED and TEMPURA using their released SGDET configurations.
            (b) Empirical transition rates across equal-count deciles of relation-query displacement magnitude $r_q$ and foreground confidence $c_q$, computed separately for each score over persistent pair adjacencies.
            Deciles are ordered from low to high score values.
            The dashed line indicates the overall transition rate of 33.7\%.
            }
            \label{fig:transition_analysis}
        \end{figure}
            
        \noindent\textbf{Backbone-controlled comparison.}\;
            We reimplement OED~\cite{wang2024oed} with the same frozen DINOv2 ViT-B/14 backbone as RTAformer to compare performance across the two frame subsets.
            The baseline's R@10 is 11.6 percentage points lower on relation-transition frames than on non-transition frames under \textit{with constraint}, highlighting the difficulty of the transition subset (Table~\ref{tab:backbone_controlled}).
            RTAformer improves R@10 over this baseline by 4.8 percentage points on transition frames, compared with 1.7 on non-transition frames.
            The same pattern holds under \textit{no constraint}, with gains of 4.1 and 1.9 percentage points, respectively.
            The larger gains on transition frames are consistent with our motivation to make temporal relation modeling more responsive to relation changes.
            RTAformer also uses fewer total and temporal-module parameters while maintaining comparable GFLOPs and inference speed.
        
            \begin{figure*}[t]
            \centering
            \includegraphics[width=1.0\textwidth]{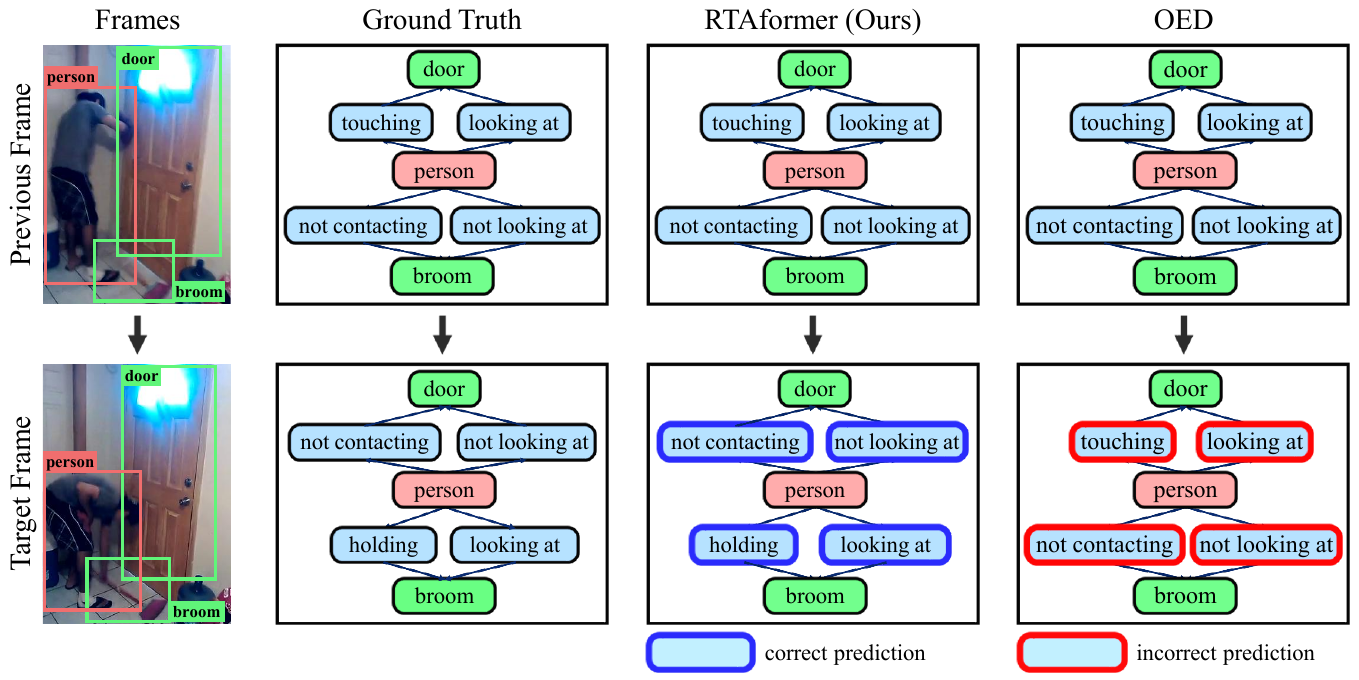}
            \caption{
            Qualitative comparison around relation transition boundaries.
            The participating entities remain visible across adjacent frames, while predicates for `person--broom' and `person--door' change.
            RTAformer updates the changed predicates, e.g., `person--holding--broom' and `person--not contacting--door', whereas OED preserves previous-frame relations such as `person--not contacting--broom' and `person--touching--door'.
            }
        \label{fig:qualitative_transition}
        \end{figure*}

        \noindent\textbf{Comparison with released baseline configurations.}\;
            As a complementary evaluation, we compare RTAformer with ViT-B/14 against OED~\cite{wang2024oed} and TEMPURA~\cite{nag2023tempura} using their released SGDET configurations.
            RTAformer achieves higher recall on the relation-transition frame subset than both baselines at all evaluated cutoffs under both constraint settings (Fig.~\ref{fig:transition_analysis}(a)).
            In particular, its R@10 exceeds OED's by 8.0 and 8.6 percentage points under \textit{with constraint} and \textit{no constraint}, respectively.
            
        \noindent\textbf{Relation-query displacement as a transition cue.}\;
            We next examine whether larger changes in relation-query representations are associated with actual predicate transitions.
            We evaluate the displacement magnitude $r_q$ from Eq.~\eqref{eq:decompose} over the 105,356 persistent pair adjacencies.
            As $r_q$ increases across equal-count deciles, transition prevalence rises monotonically from 20.9\% to 55.4\%, compared with the overall rate of 33.7\% (Fig.~\ref{fig:transition_analysis}(b)).
            Transition cases also have a larger mean $r_q$ than non-transition cases (0.396 vs.\ 0.289), and using $r_q$ directly as a transition score yields an AUROC of 0.621 without fitting a classifier.
            Similar AUROCs are observed within attention, spatial, and contacting relations (0.633/0.636/0.632).
            These findings support using relation-query displacement as an informative but imperfect cue for predicate updates.
        
            For comparison, the foreground confidence $c_q$ used to gate injection in Eq.~\eqref{eq:conf_gate} is lower for transition cases than for non-transition cases (0.517 vs.\ 0.591), with an AUROC of 0.442 when larger values are treated as indicating transitions.
            
        \noindent\textbf{Qualitative Analysis at Transition Boundaries.}\;
            Fig.~\ref{fig:qualitative_transition} shows representative relation-transition cases.
            The same entities remain visible across adjacent frames, while the predicates for `person--broom' and `person--door' change.
            For example, `person--broom' changes from \emph{not contacting} to \emph{holding}, and `person--door' changes from \emph{touching} to \emph{not contacting}.
            OED preserves the previous-frame predicates at these transition boundaries, whereas RTAformer updates the relation labels according to the current frame.
            This qualitatively illustrates relational semantic inertia and complements the transition-subset results in Fig.~\ref{fig:transition_analysis}(a).

\section{Limitations}
    Although SSIA improves same-index temporal consistency and qualitative observations suggest that same-index queries can maintain consistent \emph{subject--object} grounding across adjacent frames, DETR-style decoding does not guarantee exact cross-frame query correspondence.
    Our evaluation is limited to Action Genome under the SGDET setting, and broader evaluation across VidSGG datasets and protocols would strengthen evidence for generalization.

\section{Conclusion}
\sloppy

    Reliable video scene graph generation requires models to recognize when a previously valid relation no longer describes the observed interaction.
    We examine this challenge through \emph{relational semantic inertia}, drawing attention to outdated predicate predictions that persist even as relationships between visible entities change.
    RTAformer uses calibrated relation-query changes to guide predicate updates, and experiments on Action Genome show improved SGDET performance, with a backbone-controlled comparison revealing larger gains on relation-transition frames than on non-transition frames.
    Our analyses connect these performance observations with evidence that relation-query displacement carries information about predicate transitions.
    Together, the characterization of this failure mode and the transition-focused evaluation provide a concrete basis for future work to examine responsiveness to relation changes alongside overall prediction accuracy.


\section*{Acknowledgements}


This work was partly supported by the Institute of Information \& Communications Technology Planning \& Evaluation (IITP) under the Leading Generative AI Human Resources Development (IITP-2026-RS-2026-25544647, 40\%) grant funded by the Korean government (MSIT),
the National Research Foundation of Korea (NRF) grant funded by the Korea government (MSIT) (RS-2025-00553785, 30\%), 
and the Technology Innovation Program (K-On-Device AI Semiconductor Technology Development) (RS-2026-25617922, 30\%, Development of On-Device AI Solution and Service Technology Utilizing Task-Specialized Foundation Model for Next-Generation Cobot) funded by the Ministry of Trade, Industry and Resources (MOTIR, Korea).

\bibliographystyle{splncs04}
\bibliography{references}

\begin{thebibliography}{10}
\providecommand{\url}[1]{\texttt{#1}}
\providecommand{\urlprefix}{URL }
\providecommand{\doi}[1]{https://doi.org/#1}

\bibitem{carion2020detr}
Carion, N., Massa, F., Synnaeve, G., Usunier, N., Kirillov, A., Zagoruyko, S.: End-to-end object detection with transformers. In: ECCV. pp. 213--229 (2020). \doi{10.1007/978-3-030-58452-8_13}

\bibitem{chen2025diffvsgg}
Chen, M., Li, L., Wang, W., Yang, Y.: {DIFFVSGG}: Diffusion-driven online video scene graph generation. In: CVPR. pp. 29161--29172 (2025)

\bibitem{cherian2022spatiotemporal}
Cherian, A., Hori, C., Marks, T.K., {Le Roux}, J.: (2.5+1){D} spatio-temporal scene graphs for video question answering. In: AAAI. vol.~36, pp. 444--453 (2022)

\bibitem{cong2021sttran}
Cong, Y., Liao, W., Ackermann, H., Rosenhahn, B., Yang, M.Y.: Spatial-temporal transformer for dynamic scene graph generation. In: ICCV. pp. 16372--16382 (2021)

\bibitem{cong2023reltr}
Cong, Y., Yang, M.Y., Rosenhahn, B.: {RelTR}: Relation transformer for scene graph generation. IEEE TPAMI  \textbf{45}(9),  11169--11183 (2023). \doi{10.1109/TPAMI.2023.3268066}

\bibitem{cui2023faq}
Cui, Y.: Feature aggregated queries for transformer-based video object detectors. In: Proceedings of the IEEE/CVF Conference on Computer Vision and Pattern Recognition (CVPR). pp. 6365--6376 (2023)

\bibitem{dosovitskiy2020vit}
Dosovitskiy, A., Beyer, L., Kolesnikov, A., Weissenborn, D., Zhai, X., Unterthiner, T., Dehghani, M., Minderer, M., Heigold, G., Gelly, S., Uszkoreit, J., Houlsby, N.: An image is worth 16x16 words: Transformers for image recognition at scale. In: ICLR (2021)

\bibitem{feng2023dsgdetr}
Feng, S., Mostafa, H., Nassar, M., Majumdar, S., Tripathi, S.: Exploiting long-term dependencies for generating dynamic scene graphs. In: WACV. pp. 5130--5139 (2023)

\bibitem{huang2025esca}
Huang, J., Sethi, A., Kuo, M., Keoliya, M., Velingker, N., Jung, J., Lim, S.N., Li, Z., Naik, M.: {ESCA}: Contextualizing embodied agents via scene-graph generation. In: NeurIPS (2025)

\bibitem{im2024egtr}
Im, J., Nam, J., Park, N., Lee, H., Park, S.: {EGTR}: Extracting graph from transformer for scene graph generation. In: CVPR. pp. 24229--24238 (2024)

\bibitem{ji2020action}
Ji, J., Krishna, R., Fei-Fei, L., Niebles, J.C.: Action genome: Actions as compositions of spatio-temporal scene graphs. In: CVPR. pp. 10236--10247 (2020)

\bibitem{le2026uno}
Le, H., Chung, N., Kieu, T., Yang, J., Le, N.: {UNO}: Unifying one-stage video scene graph generation via object-centric visual representation learning. In: WACV (2026)

\bibitem{li2024sgtrplus}
Li, R., Zhang, S., He, X.: {SGTR+}: End-to-end scene graph generation with transformer. IEEE TPAMI  \textbf{46}(4),  2191--2205 (2024). \doi{10.1109/TPAMI.2023.3332246}

\bibitem{li2022apt}
Li, Y., Yang, X., Xu, C.: Dynamic scene graph generation via anticipatory pre-training. In: CVPR. pp. 13874--13883 (2022)

\bibitem{lin2024td2}
Lin, X., Shi, C., Zhan, Y., Yang, Z., Wu, Y., Tao, D.: {TD$^2$-Net}: Toward denoising and debiasing for video scene graph generation. In: Proceedings of the AAAI Conference on Artificial Intelligence. vol.~38, pp. 3495--3503 (2024). \doi{10.1609/aaai.v38i4.28137}

\bibitem{liu2022dabdetr}
Liu, S., Li, F., Zhang, H., Yang, X., Qi, X., Su, H., Zhu, J., Zhang, L.: {DAB-DETR}: Dynamic anchor boxes are better queries for {DETR}. In: ICLR (2022)

\bibitem{manasyan2025slotcontrast}
Manasyan, A., Seitzer, M., Radovic, F., Martius, G., Zadaianchuk, A.: Temporally consistent object-centric learning by contrasting slots. In: Proceedings of the Computer Vision and Pattern Recognition Conference (CVPR). pp. 5401--5411 (June 2025)

\bibitem{meinhardt2022trackformer}
Meinhardt, T., Kirillov, A., Leal-Taix{\'e}, L., Feichtenhofer, C.: {TrackFormer}: Multi-object tracking with transformers. In: CVPR. pp. 8844--8854 (2022)

\bibitem{meng2021conddetr}
Meng, D., Chen, X., Fan, Z., Zeng, G., Li, H., Yuan, Y., Sun, L., Wang, J.: Conditional {DETR} for fast training convergence. In: ICCV. pp. 3651--3660 (2021)

\bibitem{nag2023tempura}
Nag, S., Min, K., Tripathi, S., Roy-Chowdhury, A.K.: Unbiased scene graph generation in videos. In: CVPR. pp. 22803--22813 (2023)

\bibitem{nguyen2025hyperglm}
Nguyen, T.T., Nguyen, P., Cothren, J., Yilmaz, A., Luu, K.: {HyperGLM}: Hypergraph for video scene graph generation and anticipation. In: CVPR. pp. 29150--29160 (2025)

\bibitem{oord2018cpc}
van~den Oord, A., Li, Y., Vinyals, O.: Representation learning with contrastive predictive coding. arXiv preprint arXiv:1807.03748  (2018)

\bibitem{oquab2023dinov2}
Oquab, M., Darcet, T., Moutakanni, T., Vo, H.V., Szafraniec, M., Khalidov, V., Fernandez, P., Haziza, D., Massa, F., El-Nouby, A., et~al.: {DINOv2}: Learning robust visual features without supervision. TMLR  (2024)

\bibitem{rosinol2020dynamic}
Rosinol, A., Gupta, A., Abate, M., Shi, J., Carlone, L.: {3D} dynamic scene graphs: Actionable spatial perception with places, objects, and humans. In: Robotics: Science and Systems (RSS) (2020). \doi{10.15607/RSS.2020.XVI.079}

\bibitem{tang2019vctree}
Tang, K., Zhang, H., Wu, B., Luo, W., Liu, W.: Learning to compose dynamic tree structures for visual contexts. In: CVPR. pp. 6619--6628 (2019)

\bibitem{teng2021trace}
Teng, Y., Wang, L., Li, Z., Wu, G.: Target adaptive context aggregation for video scene graph generation. In: ICCV. pp. 13688--13697 (2021)

\bibitem{urooj2023shgvqa}
Urooj, A., Kuehne, H., Wu, B., Chheu, K., Bousselham, W., Gan, C., Lobo, N., Shah, M.: Learning situation hyper-graphs for video question answering. In: CVPR. pp. 14879--14889 (2023)

\bibitem{vaswani2017attention}
Vaswani, A., Shazeer, N., Parmar, N., Uszkoreit, J., Jones, L., Gomez, A.N., Kaiser, {\L}., Polosukhin, I.: Attention is all you need. In: NeurIPS. pp. 5998--6008 (2017)

\bibitem{wang2024oed}
Wang, G., Li, Z., Chen, Q., Liu, Y.: {OED}: Towards one-stage end-to-end dynamic scene graph generation. In: CVPR. pp. 27938--27947 (2024)

\bibitem{wang2023tr2}
Wang, J., Huang, J., Zhang, C., Deng, Z.: Cross-modality time-variant relation learning for generating dynamic scene graphs. In: Proceedings of the IEEE International Conference on Robotics and Automation (ICRA) (2023)

\bibitem{wang2025arn}
Wang, L., Zhang, Y., Lu, T., Liu, F., Zhang, H., Ma, J., Zhou, H.: End-to-end entity-predicate association reasoning for dynamic scene graph generation. In: ICCV. pp. 17729--17738 (2025)

\bibitem{wang2022tpi}
Wang, S., Gao, L., Lyu, X., Guo, Y., Zeng, P., Song, J.: Dynamic scene graph generation via temporal prior inference. In: ACM MM. pp. 5793--5801 (2022)

\bibitem{yang2018graphrcnn}
Yang, J., Lu, J., Lee, S., Batra, D., Parikh, D.: Graph {R-CNN} for scene graph generation. In: ECCV. pp. 670--685 (2018)

\bibitem{yin2024sgnav}
Yin, H., Xu, X., Wu, Z., Zhou, J., Lu, J.: {SG-Nav}: Online {3D} scene graph prompting for {LLM}-based zero-shot object navigation. In: NeurIPS. vol.~37, pp. 5285--5307 (2024)

\bibitem{zellers2018motifs}
Zellers, R., Yatskar, M., Thomson, S., Choi, Y.: Neural motifs: Scene graph parsing with global context. In: CVPR. pp. 5831--5840 (2018)

\bibitem{zeng2022motr}
Zeng, F., Dong, B., Zhang, Y., Wang, T., Zhang, X., Wei, Y.: {MOTR}: End-to-end multiple-object tracking with transformer. In: ECCV. pp. 659--675 (2022)

\bibitem{zhang2023dino}
Zhang, H., Li, F., Liu, S., Zhang, L., Su, H., Zhu, J., Ni, L., Shum, H.Y.: {DINO}: {DETR} with improved denoising anchor boxes for end-to-end object detection. In: ICLR (2023)

\bibitem{zhang2024tpt}
Zhang, Y., Pan, Y., Yao, T., Huang, R., Mei, T., Chen, C.W.: End-to-end video scene graph generation with temporal propagation transformer. IEEE TMM  \textbf{26},  1613--1625 (2024). \doi{10.1109/TMM.2023.3283879}

\bibitem{zhuang2025stabile}
Zhuang, W., Dong, B., Zhu, Z., Li, Z., Liu, J., Wang, Y., Hong, X., Li, X., Zuo, W.: Spatial-temporal saliency guided unbiased contrastive learning for video scene graph generation. IEEE TMM  \textbf{27},  3092--3104 (2025). \doi{10.1109/TMM.2025.3557688}

\end{thebibliography}


\begin{thebibliography}{10}
\providecommand{\url}[1]{\texttt{#1}}
\providecommand{\urlprefix}{URL }
\providecommand{\doi}[1]{https://doi.org/#1}

\bibitem{carion2020detr}
Carion, N., Massa, F., Synnaeve, G., Usunier, N., Kirillov, A., Zagoruyko, S.: End-to-end object detection with transformers. In: ECCV. pp. 213--229 (2020). \doi{10.1007/978-3-030-58452-8_13}

\bibitem{chen2025diffvsgg}
Chen, M., Li, L., Wang, W., Yang, Y.: {DIFFVSGG}: Diffusion-driven online video scene graph generation. In: CVPR. pp. 29161--29172 (2025)

\bibitem{dosovitskiy2020vit}
Dosovitskiy, A., Beyer, L., Kolesnikov, A., Weissenborn, D., Zhai, X., Unterthiner, T., Dehghani, M., Minderer, M., Heigold, G., Gelly, S., Uszkoreit, J., Houlsby, N.: An image is worth 16x16 words: Transformers for image recognition at scale. In: ICLR (2021)

\bibitem{ji2020action}
Ji, J., Krishna, R., Fei-Fei, L., Niebles, J.C.: Action genome: Actions as compositions of spatio-temporal scene graphs. In: CVPR. pp. 10236--10247 (2020)

\bibitem{le2026uno}
Le, H., Chung, N., Kieu, T., Yang, J., Le, N.: {UNO}: Unifying one-stage video scene graph generation via object-centric visual representation learning. In: WACV (2026)

\bibitem{nag2023tempura}
Nag, S., Min, K., Tripathi, S., Roy-Chowdhury, A.K.: Unbiased scene graph generation in videos. In: CVPR. pp. 22803--22813 (2023)

\bibitem{oquab2023dinov2}
Oquab, M., Darcet, T., Moutakanni, T., Vo, H.V., Szafraniec, M., Khalidov, V., Fernandez, P., Haziza, D., Massa, F., El-Nouby, A., et~al.: {DINOv2}: Learning robust visual features without supervision. TMLR  (2024)

\bibitem{vaswani2017attention}
Vaswani, A., Shazeer, N., Parmar, N., Uszkoreit, J., Jones, L., Gomez, A.N., Kaiser, {\L}., Polosukhin, I.: Attention is all you need. In: NeurIPS. pp. 5998--6008 (2017)

\bibitem{wang2024oed}
Wang, G., Li, Z., Chen, Q., Liu, Y.: {OED}: Towards one-stage end-to-end dynamic scene graph generation. In: CVPR. pp. 27938--27947 (2024)

\bibitem{wang2025arn}
Wang, L., Zhang, Y., Lu, T., Liu, F., Zhang, H., Ma, J., Zhou, H.: End-to-end entity-predicate association reasoning for dynamic scene graph generation. In: ICCV. pp. 17729--17738 (2025)

\end{thebibliography}

\end{document}


\title{Relation-Transition-Aware Transformer\\for Video Scene Graph Generation\\\large (Supplementary Material)}

\titlerunning{RTAformer: Supplementary Material}

\author{}
\authorrunning{}
\institute{}

\maketitle
\vspace{-1.15cm}

\setcounter{section}{0}
\setcounter{subsection}{0}
\setcounter{figure}{0}
\setcounter{table}{0}
\setcounter{equation}{0}

\renewcommand{\thesection}{S\arabic{section}}
\renewcommand{\thesubsection}{S\arabic{section}.\arabic{subsection}}
\renewcommand{\thefigure}{S\arabic{figure}}
\renewcommand{\thetable}{S\arabic{table}}
\renewcommand{\theequation}{S\arabic{equation}}

\section{Additional Implementation and Evaluation Details}
\label{sec:supp_impl}

In this section, we provide the implementation details omitted from the main
paper due to space constraints.

\subsection{Architecture Details}
\label{sec:supp_arch}

\begin{table}[ht]
    \centering
    \caption{Architecture configuration of RTAformer.}
    \label{tab:arch_config}
    \resizebox{\linewidth}{!}{
    \begin{tabular}{ll}
    \toprule
    Component & Configuration \\
    \midrule
    Visual backbone & Frozen DINOv2~\cite{oquab2023dinov2} ViT-S/14, ViT-B/14, ViT-L/14 \\
    Input resolution & $518 \times 518$ \\
    Transformer encoder & 6 layers \\
    Instance-pair decoder & 6 layers \\
    Relation context decoder & 6 layers \\
    Temporal aggregation decoder & 1 layer \\  
    Temporal aggregation depth $N_d$ & 3 blocks per DTA stream \\
    Hidden dimension $d_{\mathrm{model}}$ & 256 \\
    Feed-forward dimension $d_{\mathrm{ff}}$ & 2048 \\
    Number of attention heads & 8 \\
    Dropout rate & 0.1 \\
    Number of triplet queries $N_q$ & 100 \\
    \bottomrule
    \end{tabular}
    }
\end{table}

\noindent Following the recent VidSGG baseline UNO~\cite{le2026uno}, we adopt the same frozen DINOv2~\cite{oquab2023dinov2} visual backbones under the ViT-based~\cite{dosovitskiy2020vit} setting. 
Keeping the backbone frozen and identical across all compared variants ensures that performance differences arise from the model design rather than from backbone adaptation, enabling a controlled comparison under a shared visual representation.
For the Transformer modules, we adopt a standard DETR-style configuration with $d_{\mathrm{model}}=256$, $d_{\mathrm{ff}}=2048$, 8 attention heads, and $N_q=100$ triplet queries~\cite{carion2020detr}.
Here, $N_d$ denotes the number of stacked Temporal Aggregation Decoder blocks in each DTA stream. 
Each block is a single standard Transformer decoder layer consisting of self-attention, temporal cross-attention, and a feed-forward network~\cite{vaswani2017attention}.

\subsection{Training and Evaluation Details}
\label{sec:supp_training}

Training is performed in two stages.
In the first stage, we train the frame-wise spatial parser with frame-level SGG supervision and the SSIA objective $\mathcal{L}_{\mathrm{SSIA}}$.
In the second stage, we initialize from the trained spatial checkpoint, freeze both the visual backbone and the spatial parser, and optimize only the DTA module and temporal prediction heads on sequence-level clips.
Thus, the SSIA loss is applied only during spatial module training, while the temporal training stage focuses on learning transition-aware temporal aggregation.

\begin{table}[ht]
\centering
\caption{
Training details. All RTAformer training runs use a fixed random seed of 42.
}
\label{tab:training_details}
\resizebox{\linewidth}{!}{
\begin{tabular}{lcc}
\toprule
Hyperparameter & Spatial module training & Temporal module training \\
\midrule
Trainable modules
& Spatial parser + SSIA
& DTA + temporal heads \\
Frozen modules
& Visual backbone
& Visual backbone + spatial parser \\
Per-GPU batch size
& 20
& 32 \\
Number of GPUs
& 4 NVIDIA RTX A6000
& 4 NVIDIA RTX A6000 \\
Epochs
& 31
& 10 \\
Optimizer
& AdamW
& AdamW \\
Weight decay
& $1\times10^{-4}$
& $1\times10^{-4}$ \\
Base learning rate
& $1\times10^{-4}$
& $1\times10^{-4}$ \\
LR schedule
& Step
& Step \\
LR decay epoch
& 27
& 7 \\
Random seed
& 42
& 42 \\
SSIA IoU validity threshold $\tau$
& 0.3
& -- \\
Stability constant $\epsilon$
& --
& $1\times10^{-6}$ \\
TADI maximum injection strength $\gamma_{\max}$ & -- & 0.2 \\
Relation-stream delta saliency weight $\eta$ & -- & 0.05 \\

\bottomrule
\end{tabular}
}
\end{table}

\begin{table}[ht]
\centering
\caption{Loss coefficients.}
\label{tab:loss_coefficients}
\begin{tabular}{lc}
\toprule
Loss term / hyperparameter & Coefficient \\
\midrule
Object classification loss $\lambda_{\mathrm{obj}}$ & 1.0 \\
Predicate classification loss $\lambda_{\mathrm{rel}}$ & 2.0 \\
Box $\ell_1$ loss $\lambda_{\ell_1}$ & 2.5 \\
Box GIoU loss $\lambda_{\mathrm{giou}}$ & 1.0 \\
SSIA loss $\lambda_{\mathrm{SSIA}}$ & 0.1 \\
SSIA temperature $\kappa$ & 0.1 \\
\bottomrule
\end{tabular}
\end{table}

\noindent For SSIA loss, the IoU validity threshold $\tau=0.3$ in Eq.~(9) of the main paper is used only for mining valid positive anchors in the SSIA objective, not for SGDET evaluation. We use a relatively relaxed threshold because Action Genome annotates sparsely sampled frames rather than every video frame~\cite{ji2020action}. 
Thus, two adjacent annotated frames may correspond to a non-trivial temporal gap in the original video, and a persistent object can undergo substantial spatial displacement. 
To avoid discarding such valid cross-frame anchors, we combine this IoU threshold with the category-consistency constraint in Eq.~(9) of the main paper.

\subsection{Relation-Transition Subset Construction}
\label{sec:supp_transition_subset}

This section provides additional details on the construction of the relation-transition subset used in the main paper's relation-transition frame subset evaluation (Sec.~4.3 and Fig.~4).
This analysis is designed to evaluate model robustness under challenging relation-transition cases, where the same subject--object pair remains visible across adjacent annotated frames but its predicate label changes.
We therefore construct an evaluation-only subset from the Action Genome test split using ground-truth annotations. 
For each video, we sort annotated frames by frame index and consider adjacent annotated frames $(t-1,t)$, which are not necessarily consecutive frames in the original video.

\begin{table}[ht]
    \centering
    \caption{
    Statistics of the relation-transition subset used for evaluation on the Action
    Genome test split. Frame-level counts denote unique target frames containing
    at least one transition pair, while pair-adjacency-level counts denote individual
    persistent subject--object pairs matched across adjacent annotated frames.
    }
    \label{tab:transition_subset_stats}
    \begin{tabular}{lr}
    \toprule
    Quantity & Count \\
    \midrule
    Test videos & 1,750 \\
    Annotated test frames & 56,880 \\
    Relation-transition event frames & 25,883 \\
    \quad Attention-family transition frames & 16,567 \\
    \quad Spatial-family transition frames & 7,177 \\
    \quad Contacting-family transition frames & 11,835 \\
    \midrule
    Persistent subject--object pair-adjacencies & 105,356 \\
    Transition pair-adjacencies & 35,523 \\
    \quad Attention-family transition pair-adjacencies & 20,349 \\
    \quad Spatial-family transition pair-adjacencies & 7,878 \\
    \quad Contacting-family transition pair-adjacencies & 14,093 \\
    \bottomrule
    \end{tabular}
\end{table}

\noindent\textbf{Persistent subject--object pairs.}
Since Action Genome provides person-centric triplet annotations, we use the annotated person box as the subject and non-person boxes as objects.
A subject--object pair-adjacency is retained only if the subject box persists
with IoU $> 0.3$ and an object of the same category is matched across the two
frames by class-wise IoU matching with IoU $> 0.3$.

\noindent\textbf{Transition criterion.}
For each persistent subject--object pair $p=(s,o)$ matched between frames
$t-1$ and $t$, let $\mathcal{R}_{t}^{f}(p)$ denote its predicate label set for
relation family $f \in \{\mathrm{attention}, \mathrm{spatial}, \mathrm{contacting}\}$
at frame $t$. We count a family-wise transition when the predicate sets in the
two frames differ and have no overlap:
\begin{equation}
    I_{\mathrm{trans}}^{f}(p,t)
    =
    \ind\!\bigl[
    \mathcal{R}_{t-1}^{f}(p) \neq \mathcal{R}_{t}^{f}(p)
    \;\wedge\;
    \mathcal{R}_{t-1}^{f}(p) \cap \mathcal{R}_{t}^{f}(p) = \varnothing
    \bigr].
\end{equation}
A persistent pair $p$ is included in the transition subset if
$I_{\mathrm{trans}}^{f}(p,t)=1$ for at least one relation family $f$.
Partially overlapping predicate sets are excluded from the transition subset, i.e., not counted as transition cases, to construct a conservative subset with unambiguous multi-label changes.
For frame-level evaluation, we consider only target frames containing at
least one persistent subject--object pair with the preceding annotated frame.
A target frame belongs to the relation-transition subset if at least one
persistent pair satisfies the transition criterion in any relation family,
and to the non-transition subset otherwise.
These subsets partition the 47,357 eligible target frames into 25,883
relation-transition frames and 21,474 non-transition frames.
Non-transition therefore denotes the absence of transitions under our
criterion, rather than necessarily unchanged predicate sets.
Pair-adjacency-level statistics count individual persistent subject--object
pairs across adjacent annotated frames.

\noindent\textbf{Scope of baseline comparison.}
The relation-transition subset is newly constructed by filtering the Action Genome test split for persistent subject--object pairs with predicate changes.
As a result, previously reported full-test-set results are not directly comparable to this subset evaluation, and baselines must be re-evaluated under the same subset construction and evaluation protocol.
Evaluating a method on this subset requires its pretrained checkpoint, exact evaluation hyperparameters, and a compatible SGDET evaluation pipeline. 
We therefore report transition-subset results only for TEMPURA~\cite{nag2023tempura} and OED~\cite{wang2024oed}, for which compatible public resources allowed controlled evaluation. 
For UNO~\cite{le2026uno}, DIFFVSGG~\cite{chen2025diffvsgg}, and ARN~\cite{wang2025arn}, we were unable to obtain the combination of pretrained checkpoints, exact evaluation hyperparameters, and compatible SGDET evaluation code required for controlled subset evaluation at the time of our experiments.

\section{Additional Analysis of Spatio-Semantic Instance Alignment}
\sloppy
\label{sec:supp_ssia}

    We first clarify the index correspondence used throughout this analysis.
    RTAformer represents each candidate interaction by a triplet query indexed by $q \in \{1,\dots,N_q\}$. 
    At frame $t$, the instance-pair query $Q^{ins}_{t,q}$ predicts the subject and object of the $q$-th candidate, while the relation query $Q^{rel}_{t,q}$ predicts the predicate for the same candidate pair. 
    This correspondence follows from the cascaded decoding design: $Q^{rel}_{t}$ is produced by applying the Relation Context Decoder to $Q^{ins}_{t}$, with no sorting, reassignment, or re-indexing between the instance-pair and relation stages. 
    Therefore, the predictions at index $q$ are combined to form the $q$-th triplet $\langle \mathrm{subject}, \mathrm{predicate}, \mathrm{object}\rangle_q$. 
    During training, Hungarian matching supervises the subject, object, and predicate predictions at the assigned query index as a single triplet, using a joint cost over object classification, predicate classification, and box localization.
    Thus, within each frame, index $q$ defines one triplet candidate shared by the instance-pair and relation predictions. 
    
    SSIA is designed to stabilize this index-level instance-pair grounding across adjacent frames.
    This property is important for RTAformer because TADI computes same-index relation-query displacements between adjacent frames. 
    If the same-index queries were not grounded to consistent subject--object pairs, the resulting displacement could reflect representational drift rather than a genuine relation transition.
    Therefore, this section provides additional analysis showing that SSIA induces temporally consistent instance-pair query representations.
    
    \subsection{Illustration of SSIA Validity Masking}
    \label{sec:supp_ssia_masking}

    \noindent Before analyzing the trained query representations, we first illustrate the validity masking used in SSIA. 
    SSIA does not force every same-index candidate pair $(Q^{ins}_{t,q_m}, Q^{ins}_{t-1,q_m})$ constructed from a GT-matched target query to be close. 
    Instead, a candidate pair is used as a positive anchor only when the target-frame object matched to query $q_m$ has a same-category object with sufficient spatial overlap in the previous frame. 
    Otherwise, the pair is masked out from the SSIA contrastive objective.
    
    This masking prevents SSIA from incorrectly pulling together query embeddings when the corresponding object-side entity is not verified to persist across adjacent annotated frames.
    Thus, SSIA encourages temporal consistency only for reliable cross-frame anchors, rather than imposing same-index alignment unconditionally.

    \begin{figure}[ht]
        \centering
        \includegraphics[width=0.45\linewidth]{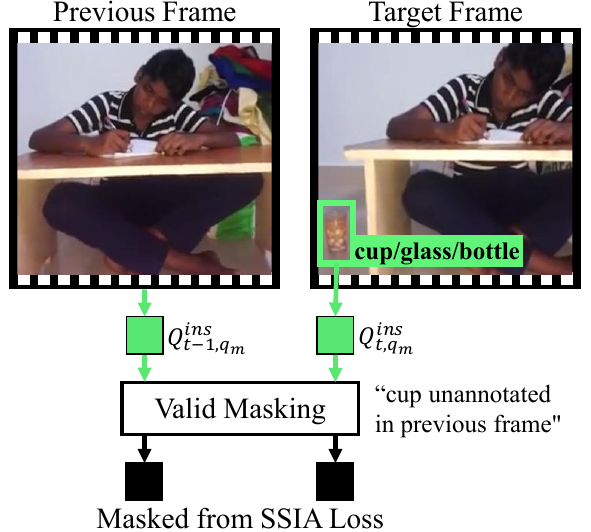}
        \caption{
        Example of SSIA validity masking. The target frame contains an annotated
        \emph{cup/glass/bottle}, while the corresponding object is not annotated in
        the previous frame. Therefore, the same-index pair
        $(Q^{ins}_{t,q_m}, Q^{ins}_{t-1,q_m})$ fails the spatio-semantic validity check
        and is not used as a positive anchor in the SSIA loss.
        }
        \label{fig:supp_ssia_masking}
    \end{figure}

    \subsection{Representation-Space Visualization of Same-Index Query Consistency}
    \label{sec:supp_ssia_tsne}

    \begin{figure}[t]
        \centering
        \includegraphics[width=\linewidth]{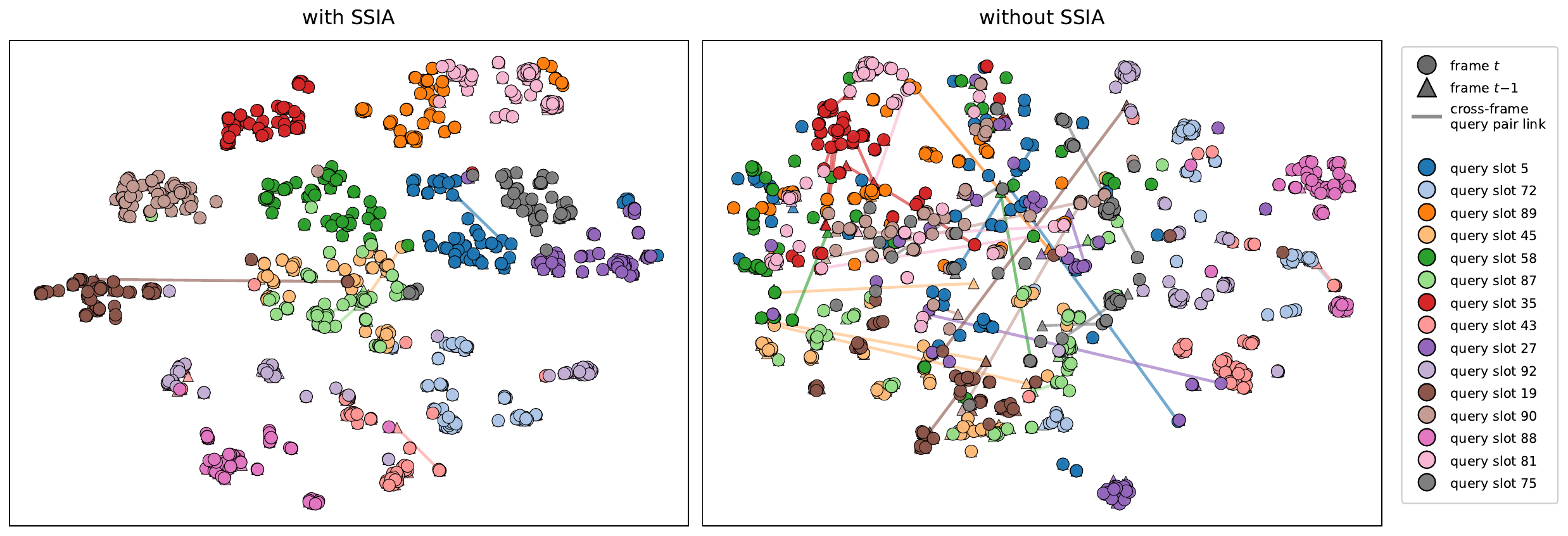}
        \caption{
        t-SNE visualization of cross-frame instance-pair query representations.
        Circles and triangles denote embeddings from frame $t$ and frame $t-1$,
        respectively, and colors indicate query slot indices. A line segment connects
        the same-index temporal query pair $(Q^{ins}_{t,q}, Q^{ins}_{t-1,q})$.
        With SSIA, same-slot embeddings form more compact clusters with shorter
        cross-frame segments, indicating improved temporal consistency of
        instance-pair query anchors. Without SSIA, same-slot embeddings are more
        dispersed and connected by longer, crossing segments, suggesting stronger
        cross-frame representational drift.
        }
        \label{fig:supp_ssia_tsne}
    \end{figure}
    Using the same spatio-semantic persistence criterion for sample selection, we qualitatively examine the representation space of same-index instance-pair queries. 
    We visualize instance-pair query representations extracted from the Action Genome test split using t-SNE.
    We compare two spatial parsers trained with the same configuration, with and without SSIA. For both variants, we collect GT-matched and class-consistent cross-frame pairs whose bounding boxes satisfy IoU $>0.3$ across adjacent annotated frames. 
    We then select the 15 most represented query slots and randomly sample 60 temporal pairs per slot, yielding 900 temporal instance pairs per model.
    For each variant, embeddings from frame $t$ and frame $t-1$ are projected into a shared two-dimensional t-SNE space.

    \noindent As shown in Fig.~\ref{fig:supp_ssia_tsne}, SSIA produces a more organized
    query representation space. Same-index query embeddings from adjacent frames
    remain closer to each other, whereas the model trained without SSIA shows more
    dispersed embeddings and longer cross-frame links. This supports the role of
    SSIA as a training-time regularizer that reduces instance-level representation
    drift before relation-query displacement is used by TADI.

    \subsection{Temporal Grounding Consistency of Same-Index Queries}
    \label{sec:supp_ssia_grounding}
    
\begin{figure}[ht]
        \centering
        \includegraphics[width=\linewidth]{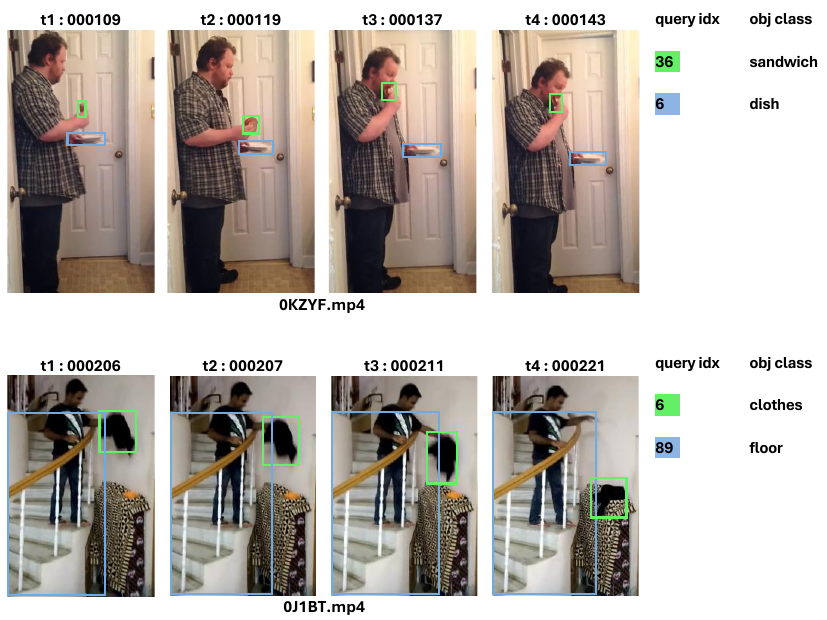}
        \caption{
        Temporal grounding of same-index queries. The SSIA-trained frame-wise parser
        is applied independently to each annotated frame, without temporal
        aggregation. Within each clip, the same color denotes the same query index.
        Person boxes are omitted for clarity.
        }
        \label{fig:supp_ssia_grounding}
    \end{figure}

    We next examine whether the representation-space alignment in
    Fig.~\ref{fig:supp_ssia_tsne} has a corresponding image-space effect.
    We use the SSIA-trained frame-wise spatial parser from Phase~1, before applying
    DTA, and run it independently on four consecutive annotated frames from each
    clip. No reference frame, temporal aggregation, tracking, IoU matching, or other
    cross-frame association is used. Thus, boxes with the same color in
    Fig.~\ref{fig:supp_ssia_grounding} are direct frame-wise predictions from the
    same query index. 
    Following the person-centric annotations of Action Genome, we omit person boxes and visualize only object-side predictions for clarity.
    Since the subject side is the person in these visualized examples, object-side consistency corresponds to stable person–object candidate grounding.
    
    Fig.~\ref{fig:supp_ssia_grounding} shows that the visualized same-index queries
    maintain object-side grounding across annotated frames under purely frame-wise
    inference. 
    In the first clip, the same query indices consistently localize the
    \emph{sandwich} and \emph{dish} despite hand motion and object displacement. 
    In the second clip, they remain grounded to \emph{clothes} and the \emph{floor}
    region across the frame sequence.
    This qualitative check complements Fig.~\ref{fig:supp_ssia_tsne} and supports the role of SSIA in stabilizing instance-pair query anchors before TADI computes same-index relation-query
    displacements.
        
\clearpage

\renewcommand{\refname}{Supplementary References}
\bibliographystyle{splncs04}
\bibliography{references}